\documentclass[10pt,twocolumn,letterpaper]{article}

\usepackage[OT1]{fontenc}
\usepackage{fola}

\usepackage{cuted}
\usepackage{amsmath,amssymb}
\usepackage{booktabs}
\usepackage{multirow}
\usepackage{xspace}
\usepackage{pifont}
\usepackage{algorithm}
\usepackage[noend]{algpseudocode}
\usepackage{colortbl}
\newcommand{\method}{FoLA\xspace}
\newcommand{\policy}{FoLA-P\xspace}
\newcommand{\und}{understanding branch\xspace}
\newcommand{\gen}{generation branch\xspace}
\newcommand{\data}{FoLA-14K\xspace}
\newcommand{\vel}{\mathbf{v}}
\newcommand{\sg}{\operatorname{sg}}
\newcommand{\up}{$\uparrow$}
\newcommand{\sig}{$^\ast$}

\usepackage{microtype}
\makeatletter
\renewcommand{\paragraph}{\@startsection{paragraph}{4}{\z@}{5pt plus 1pt minus 1pt}{-1em}{\normalfont\normalsize\bfseries}}
\makeatother

\definecolor{linkblue}{rgb}{0.21,0.49,0.74}
\usepackage[pagebackref,breaklinks,colorlinks,allcolors=linkblue]{hyperref}
\hypersetup{pdftitle={FoLA: Latent Forethought Guides Unified Image Generation},pdfauthor={Fuxiang Zhai, Sixiang Chen, Tengjun Huang, Yingjin Li, Lei Zhu}}

\title{FoLA: Latent Forethought Guides Unified Image Generation}

\author{
Fuxiang Zhai$^{1}$ \quad Sixiang Chen$^{1}$ \quad Tengjun Huang$^{1}$ \quad Yingjin Li$^{1}$ \quad Lei Zhu$^{1,2}$\\
$^{1}$The Hong Kong University of Science and Technology (Guangzhou)\\
$^{2}$The Hong Kong University of Science and Technology\\
{\tt\small fzhai598@connect.hkust-gz.edu.cn}\\
{\small Project page: \url{https://fuxiangzhai.github.io/FoLA/}}
}

\begin{document}
\maketitle
\raggedbottom
\widowpenalty=10000
\begin{strip}
\centering
\includegraphics[width=\linewidth]{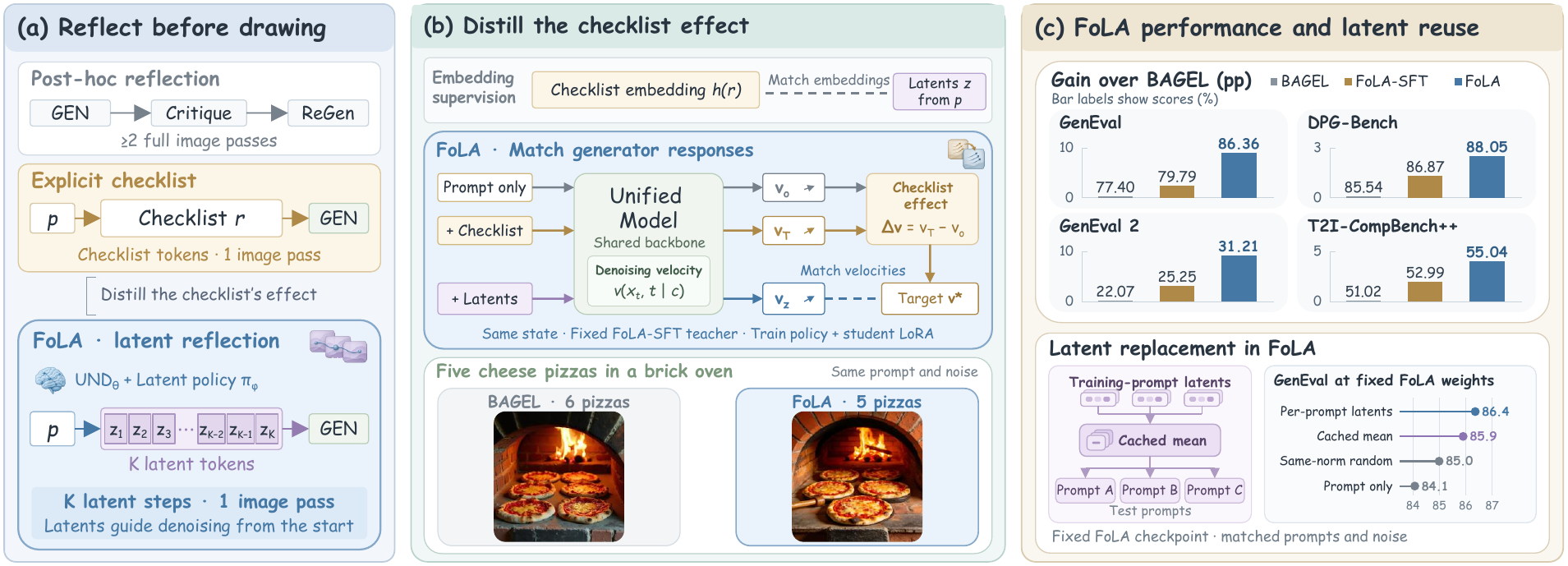}
\captionof{figure}{\textbf{FoLA distills the effect of explicit checklists into latent guidance.}
(a)~FoLA writes latent tokens from the prompt before denoising and uses them in a single generation pass.
(b)~We distill the fixed teacher's checklist-induced velocity change.
The paired example uses the same prompt and noise for five cheese pizzas in a brick oven, with rear flames and a wooden handle.
(c)~Top: bar heights show gains over BAGEL, with absolute scores labeled.
Bottom: GenEval at the fixed joint FoLA checkpoint with per-prompt latents, one cached mean latent context, same-norm random latents, or no latents (\cref{sec:joint_results}).}
\label{fig:teaser}
\end{strip}
\begin{abstract}
Unified multimodal models can identify visual requirements in a prompt that their generators still miss.
A prompt-derived checklist can help by making those requirements explicit before denoising.
However, this guidance must be decoded as additional text for every prompt, and matching the checklist's text representation with latent tokens does not transfer its effect.
We introduce \textbf{FoLA (Forethought with Latent Actions)} to internalize the checklist's effect on the generator as latent guidance.
At each sampled state on the student's denoising trajectory, a fixed teacher predicts velocities from the prompt alone and from the same prompt with the checklist appended.
We use the difference between these predictions to amplify the checklist's effect in a training target for jointly optimizing the latent policy and the student's generation LoRA.
At inference, the policy writes latent tokens once from the prompt before denoising, and the generator uses the same latent context throughout a single generation pass.
Under matched evaluation, joint FoLA improves over both the released BAGEL weights and the FoLA-SFT initialization on all four benchmarks.
Relative to BAGEL, FoLA gains 8.96 points on GenEval, 9.15 on GenEval~2, 2.51 on DPG-Bench, and 4.01 on T2I-CompBench++.
Among failures that BAGEL's own understanding branch can detect, FoLA fixes 39\%, versus 26\% for explicit checklists.
\end{abstract}

\section{Introduction}
\label{sec:intro}

Unified multimodal models (UMMs) such as BAGEL~\cite{deng2025emerging}, Janus-Pro~\cite{chen2025janus}, and Show-o~\cite{xie2025show} combine image understanding and generation in a single model.
Their generators can still miss requirements that their understanding branches can identify~\cite{niu2026does,wang2026quantifying}.
For example, drawing two red cups to the left of a blue plate requires producing the right number of objects, binding colors to the correct objects, and preserving the specified spatial relation.
Post-hoc reflection uses feedback on a draft image to guide a subsequent pass~\cite{zhuo2025reflection,li2025reflect}.
A prompt-derived checklist brings guidance into the first pass by making visually checkable requirements explicit before denoising.
T2I-R1 uses a textual plan to guide synthesis~\cite{jiang2026t2i}, and LAC shows how continuous latent actions can condition a unified generator~\cite{zhai2026latent}.

A checklist calls attention to details already specified in the prompt that may be missed as the image takes shape.
One way to transfer this guidance to latent tokens is to train them to match the checklist's hidden states, using similarity to the text representation as the objective.
To train latent tokens to reproduce the checklist's influence on generation, we supervise them through the predictions they induce in the generator.
In a flow-based model, the predicted velocity supplies the update direction for the noisy image latent at each denoising step.
Comparing velocity predictions at the same state therefore lets us measure how the checklist changes that update.
This gives latent supervision a target directly tied to how the generator uses the checklist.

\begin{figure*}[t]
\centering
\includegraphics[width=\linewidth]{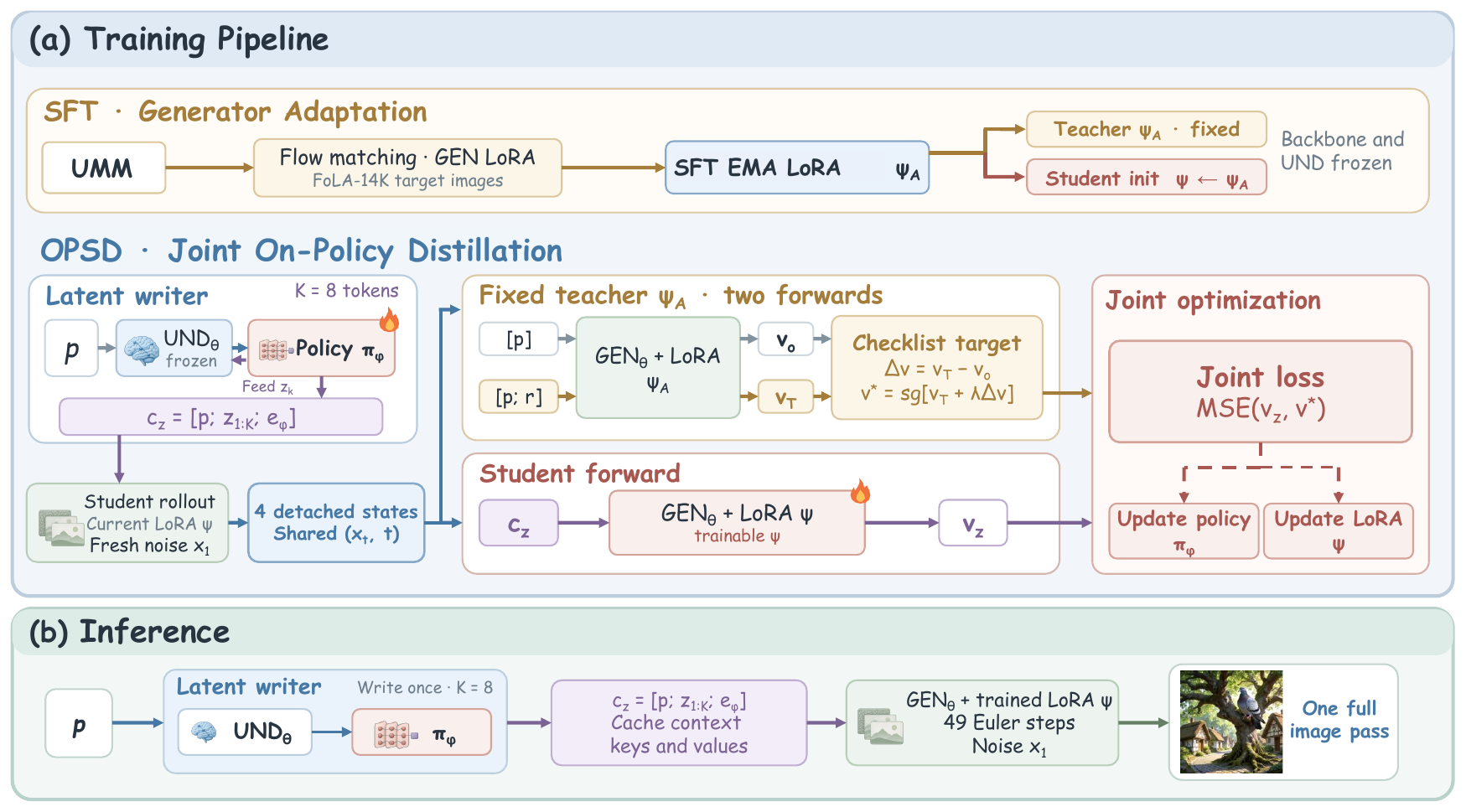}
\caption{\textbf{\method training and inference.}
\textbf{(a) Training.} SFT initializes the teacher and the student.
OPSD (on-policy self-distillation) uses states from the current
student's denoising trajectory. At each state, the fixed teacher predicts
velocities with and without the checklist. Their difference shapes the
target for jointly training the latent policy and student LoRA.
\textbf{(b) Inference.} The policy writes latent tokens once from the
prompt, and the generator reuses the latent context throughout denoising.}
\label{fig:method}
\vspace{-6pt}
\end{figure*}

We introduce \method (Forethought with Latent Actions) to internalize the checklist's effect on denoising as latent guidance (\cref{fig:method}).
Supervised fine-tuning (SFT) trains a generation LoRA with target images and checklist conditioning to initialize the teacher and student.
During on-policy self-distillation (OPSD), a latent policy uses the frozen understanding branch to write continuous tokens from the prompt, and the student generator attends to them throughout denoising.
At sampled states on the student's trajectory, the fixed teacher predicts velocities with and without the checklist.
We extrapolate the checklist-conditioned velocity along their difference to form the student's training target.
Matching this target jointly updates the latent policy through attention to the latent tokens and the generation LoRA through its velocity predictions (\cref{sec:joint_opsd}).
At inference, the policy writes the latent context once for single-pass generation.

Under matched prompts, initial noise, and evaluation, joint \method improves over released BAGEL and its FoLA-SFT initialization on GenEval, GenEval~2, DPG-Bench, and T2I-CompBench++ (\cref{tab:geneval}).
These benchmarks cover visual composition and detailed scene descriptions.
Relative to BAGEL, it gains 8.96 points on GenEval and 9.15 on GenEval~2.
With matched update budgets, joint training yields higher measured scores on all four benchmarks than training either the latent policy or the generation LoRA alone (\cref{sec:joint_results}).
At fixed joint weights, prompt-only generation retains much of the improvement, and enabling latents further improves GenEval (\cref{tab:joint_latent}).

\noindent Our main contributions are as follows:
\begin{itemize}
    \item We introduce \method to internalize the checklist's effect on denoising as latent guidance. A fixed teacher's velocity difference with and without the checklist defines the supervision target. On states visited by the student, it jointly trains the latent policy and generation LoRA.

    \item We construct \data, a training dataset of 14{,}250 prompts paired with target images and prompt-derived checklists. The checklists specify visually checkable requirements, supporting both checklist-conditioned generator adaptation and the subsequent distillation of checklist guidance into latent tokens.

    \item We show that joint \method improves over released BAGEL and the FoLA-SFT initialization on GenEval, GenEval~2, DPG-Bench, and T2I-CompBench++ under matched evaluation. Relative to BAGEL, it gains 8.96 points on GenEval and 9.15 points on GenEval~2.

    \item We find that joint training achieves higher measured scores on all four benchmarks than training either the latent policy or the generation LoRA alone under the same update budget. At the same joint checkpoint, enabling latent guidance further improves GenEval by 2.29 points.
\end{itemize}

\section{Related Work}
\label{sec:related}

\subsection{Unified Understanding and Generation}
Unified multimodal models bring visual understanding and image synthesis into one architecture.
Show-o combines autoregressive modeling and discrete diffusion in a single transformer~\cite{xie2025show}, and Janus-Pro uses separate visual encoders for understanding and generation~\cite{chen2025janus}.
BAGEL connects specialized understanding and generation experts through shared self-attention~\cite{deng2025emerging}, providing an interface through which the two branches can exchange information.

Controlled evaluations show that unified models can infer answers in text that they struggle to express in images~\cite{niu2026does,wang2026quantifying}.
This gap motivates using the understanding branch to supply conditioning for image synthesis.
In \method, a policy reads hidden states from BAGEL's frozen understanding branch and writes continuous conditioning tokens. The generation branch attends to them.

\subsection{Reflection and Planning}
Reflection methods diagnose a rendered image and use the diagnosis to guide a subsequent generation pass.
ReflectionFlow uses explicit feedback to assess and refine its earlier outputs~\cite{zhuo2025reflection}.
Reflect-DiT uses previous images and the corresponding natural-language critiques as in-context examples for the denoiser~\cite{li2025reflect}.

Planning from the prompt brings guidance into the first generation pass.
T2I-R1 produces a semantic-level textual plan before autoregressive image-token generation~\cite{jiang2026t2i}.
In \method, a prompt-derived checklist highlights visual requirements likely to be missed and supplies guidance during training.
At inference, the policy writes eight continuous tokens directly from the prompt before denoising, and the generator reuses them at every denoising step.

\subsection{Latent Reasoning and Distillation}
Continuous states can carry reasoning across steps. Coconut feeds hidden states back as input embeddings for language reasoning~\cite{hao2024training}. In visual understanding, LVR learns to reconstruct selected visual tokens~\cite{li2026latent}, and CoVT trains continuous tokens with supervision from vision experts~\cite{qin2025chain}.
For image synthesis, MILR searches continuous tokens at test time with an image-quality critic~\cite{mi2026milr}.
In BAGEL, LAC rolls out latent actions for planning, drafting, diagnosis and refinement before flow sampling. It aligns these actions to teacher-derived priors and uses terminal image rewards to optimize both latent actions and flow transitions~\cite{zhai2026latent}.

Context distillation trains a student to reproduce predictions made with additional conditioning~\cite{snell2022learning}. In diffusion models, guidance distillation matches classifier-free guided predictions~\cite{meng2023distillation}.
PE-OPSD matches enhanced-prompt predictions from an EMA teacher at states visited by a raw-prompt student~\cite{lin2026pe}.
\method evaluates a fixed FoLA-SFT teacher with and without the checklist at the same states along the student's latent-conditioned denoising trajectory.
The resulting velocity difference defines the target for the latent policy and generation LoRA (\cref{sec:velocity_target,sec:joint_opsd}).

\section{Method}
\label{sec:method}

\method learns to guide image denoising with latent tokens written from the prompt. As shown in \cref{fig:method}, SFT adapts the generation branch to use textual checklists, yielding FoLA-SFT, the shared initialization of teacher and student. OPSD holds the teacher fixed and jointly trains the latent policy and student generation LoRA. At states visited by the student, appending the checklist changes the teacher's velocity prediction. This change supplies the supervision for latent guidance. At inference, the policy writes the latent tokens once before denoising.

\subsection{Backbone and SFT Initialization}
\label{sec:sft}

\paragraph{Backbone.}
BAGEL~\cite{deng2025emerging} connects its understanding and generation branches through shared self-attention. The \gen generates images with rectified flow~\cite{lipman2022flow} in the latent space of a variational autoencoder (VAE). Given a noisy image latent $x_t$ at time $t$ and a conditioning sequence $c$, it predicts a velocity $\vel_{\theta,\psi}(x_t,t\mid c)$. Here $\theta$ denotes the frozen backbone and $\psi$ denotes the generation LoRA. The generation branch attends to text tokens and continuous embeddings through cached keys and values.

\paragraph{Training records.}
We adapt teacher records described in LAC~\cite{zhai2026latent} into \data, a set of 14{,}250 prompts with target images and prompt-derived checklists. The prompts come from ShareGPT, BLIP3o-60K and prompt sets written in the style of GenEval and WISE. For each prompt $p$, Seedream 4.5 renders the target image and Gemini 2.5 Flash writes a checklist $r$ of 3--5 visually checkable requirements likely to be missed. We remove the introductory verbs and join the checklist items into a single sentence so it can be appended to the prompt. An audit against the four evaluation benchmarks finds no exact prompt matches.

\paragraph{SFT.}
We train a rank-32 generation LoRA~\cite{hu2021lora} with flow matching on the target images. Appending $r$ to $p$ for 70\% of the prompts exposes the model to checklist-conditioned and prompt-only generation. We take the exponential moving average (EMA) of the LoRA weights after six epochs, denoted $\psi_A$; we call the resulting model FoLA-SFT. This adapter stays fixed in the OPSD teacher and initializes the student's trainable adapter $\psi$.

\subsection{Checklist-Conditioned Velocity Target}
\label{sec:velocity_target}

We evaluate the fixed FoLA-SFT teacher twice at each sampled student state $(x_t,t)$, once with $p$ and once with $c_T=[p;\,r]$,
\begin{equation}
\begin{aligned}
\vel_o &= \vel_{\theta,\psi_A}(x_t,t\mid p),\\
\vel_T &= \vel_{\theta,\psi_A}(x_t,t\mid c_T).
\end{aligned}
\end{equation}
The teacher weights, time and noisy image latent are identical in both evaluations. The velocity difference $\vel_T-\vel_o$ therefore measures the change in the predicted denoising direction due to the checklist at that state. We construct the target by extrapolating from the checklist-conditioned prediction along this difference,
\begin{equation}
\vel^\star = \sg\!\left[\vel_T + \lambda(\vel_T-\vel_o)\right].
\label{eq:teacher_target}
\end{equation}
We use $\lambda=1$. The operator $\sg$ stops gradients through the target.
For $\lambda=0$ the student imitates the checklist-conditioned teacher; for $\lambda>0$ the target moves further along the checklist direction, analogous to classifier-free guidance with the checklist as the condition and the prompt as the reference.
Unlike guidance, this extrapolation is applied only to the training target, so inference needs neither the checklist nor a second forward pass.
Because the target is defined in velocity space, it supervises what the checklist does to the denoising update rather than how the checklist is represented, which is the property our hidden-state alignment baseline lacks (\cref{sec:fixed_generator}).

\subsection{Prompt-Conditioned Latent Forethought}
\label{sec:latent_forethought}

The frozen \und reads a short cue, ``An image is being generated for the prompt: $p$. Decide what the finished image must get right to match the prompt exactly.'' Let $a(p)$ denote this cue and $h_0$ its final hidden state. At step $k$, the policy $\pi_\phi$ maps the previous hidden state to a continuous embedding $z_k$. Feeding that embedding back into the \und yields the next hidden state~\cite{hao2024training},
\begin{equation}
z_k = \sigma\,\pi_\phi(h_{k-1}), \qquad
h_k = \mathrm{UND}_\theta([a(p);\,z_{1:k}]).
\label{eq:rollout}
\end{equation}
Here $\mathrm{UND}_\theta$ returns the final hidden state, using a cache for the cue and preceding latent inputs. We use $\sigma=0.02$ and $K=8$ steps. The generator receives the original prompt followed by the eight policy outputs and a learned end embedding $e_\phi$,
\begin{equation}
c_z = [p;\,z_{1:K};\,e_\phi].
\end{equation}
The policy parameters $\phi$ include $e_\phi$; the policy is an MLP with two 2{,}048-unit hidden layers, GELU and LayerNorm, mapping 3{,}584-dimensional hidden states to embeddings (26M stored parameters). The tokens in $c_z$ condition the denoising of $x_t$.
Writing the latents with the frozen understanding branch, rather than learning free embeddings, lets the policy reuse the branch's reading of the prompt.

\begin{table*}[t]
\caption{\textbf{Comparison with state-of-the-art generators.} The upper blocks report published scores under their own protocols~\cite{cai2026hidream,diao2026sensenova}; $^\ddagger$ marks third-party evaluations~\cite{cai2026hidream,chen2026llada}. The last block uses our matched prompts and initial noise, under which released BAGEL scores 77.40 on GenEval rather than its reported 82.0. GenEval~2: Soft-TIFA GM; T2I-C++: mean of eight T2I-CompBench++ categories. Params excludes text encoders and VAEs (A: activated). Bold: best per column; underline: best among BAGEL-based methods (14B, A7B). $^\ast$: paired 95\% CI of the gain over BAGEL above zero.}
\label{tab:geneval}
\centering\setlength{\tabcolsep}{4.5pt}
\resizebox{\textwidth}{!}{\begin{tabular}{@{}lllcccccccccc@{}}
\toprule
& & & \multicolumn{7}{c}{GenEval} & DPG & GenEval~2 & T2I-C++ \\
\cmidrule(lr){4-10}
Type & Method & Params & Single & Two & Count & Color & Pos. & Attr. & Overall\up & Overall\up & GM\up & Mean\up \\
\midrule
\multirow{3}{*}{\shortstack[l]{Closed-source\\generators}}
& GPT Image 2 & -- & 99.0 & 98.0 & 85.0 & \textbf{93.0} & 85.0 & 77.0 & 89.0 & 85.98 & -- & -- \\
& Nano Banana 2 & -- & \textbf{100.0} & 96.0 & 71.0 & 84.0 & 86.0 & 65.0 & 83.0 & 86.90 & -- & -- \\
& Seedream 4.5 & -- & -- & -- & -- & -- & -- & -- & -- & \textbf{88.63} & -- & -- \\
\midrule
\multirow{3}{*}{\shortstack[l]{Open\\generators}}
& FLUX.2-dev & 32B & \textbf{100.0} & \textbf{99.0} & 79.0 & \textbf{93.0} & 73.0 & 78.0 & 87.0\rlap{$^\ddagger$} & 87.57\rlap{$^\ddagger$} & -- & -- \\
& HunyuanImage-3.0~\cite{cao2025hunyuanimage} & 80B (A13B) & \textbf{100.0} & 92.0 & 48.0 & 82.0 & 42.0 & 63.0 & 72.0\rlap{$^\ddagger$} & 86.10\rlap{$^\ddagger$} & -- & -- \\
& Qwen-Image~\cite{wu2025qwen} & 20B & 99.0 & 92.0 & \textbf{89.0} & 88.0 & 76.0 & 77.0 & 87.0 & 88.32 & -- & -- \\
\midrule
\multirow{3}{*}{\shortstack[l]{Unified\\models}}
& Janus-Pro-7B~\cite{chen2025janus} & 7B & 99.0 & 89.0 & 59.0 & 90.0 & 79.0 & 66.0 & 80.0 & 84.19 & -- & -- \\
& OmniGen2~\cite{wu2025omnigen2} & 3B+4B & \textbf{100.0} & 95.0 & 64.0 & 88.0 & 55.0 & 76.0 & 80.0 & 83.57 & -- & -- \\
& TUNA~\cite{liu2026tuna} & 7B & \textbf{100.0} & 97.0 & 81.0 & 91.0 & \textbf{88.0} & \textbf{83.0} & \textbf{90.0} & 86.76 & -- & -- \\
\midrule
\multirow{4}{*}{\shortstack[l]{Explicit\\reasoning}}
& GoT~\cite{fang2025got} & 3B+2.6B & 99.0 & 69.0 & 67.0 & 85.0 & 34.0 & 27.0 & 64.0 & -- & -- & -- \\
& IRG~\cite{huang2026interleaving} & 14B (A7B) & 98.0 & 94.0 & 83.0 & 86.0 & \underline{74.0} & 73.0 & 85.0 & -- & -- & -- \\
& Uni-CoT~\cite{qin2026uni} & 14B (A7B) & 99.0 & \underline{96.0} & 84.0 & \underline{92.0} & 57.0 & 71.0 & 83.0 & -- & -- & -- \\
& Think in Strokes~\cite{zhang2026think} & 14B (A7B) & 99.0 & 94.0 & 75.0 & 87.0 & 72.0 & 63.0 & 82.0 & -- & -- & -- \\
\midrule
\multirow{2}{*}{\shortstack[l]{Ours, matched\\protocol}}
& BAGEL (reproduced) & 14B (A7B) & 98.8 & 92.4 & 77.5 & 86.7 & 48.5 & 60.5 & 77.40 & 85.54 & 22.07 & 51.02 \\
& \cellcolor{linkblue!8}\textbf{\method (ours)} & \cellcolor{linkblue!8}14B (A7B) +26M & \cellcolor{linkblue!8}\underline{\textbf{100.0}} & \cellcolor{linkblue!8}94.4 & \cellcolor{linkblue!8}\underline{84.4} & \cellcolor{linkblue!8}88.8 & \cellcolor{linkblue!8}71.0 & \cellcolor{linkblue!8}\underline{79.5} & \cellcolor{linkblue!8}\underline{86.36}\sig & \cellcolor{linkblue!8}\underline{88.05}\sig & \cellcolor{linkblue!8}\textbf{31.21}\sig & \cellcolor{linkblue!8}\textbf{55.04}\sig \\
\bottomrule
\end{tabular}}
\end{table*}

\subsection{Joint On-Policy Distillation}
\label{sec:joint_opsd}

\paragraph{State sampling.}
For each prompt, we form $c_z$ with the current policy and run an Euler rollout with the current student generator from fresh Gaussian noise. We use classifier-free guidance of 4 on a stride-3 time grid. We retain the initial state and sample one state from each of three strata of the remaining grid. The four states are on-policy because they come from the student's own latent-conditioned trajectory. We detach the states before evaluating the teacher and student.
Sampling states from the student itself matters because the latents change the trajectory: states from a prompt-only or teacher rollout would supervise the student where it does not go at inference, an exposure bias that on-policy \mbox{distillation}~avoids~\cite{lin2026pe}.

\paragraph{Joint objective.}
At each retained state, the student predicts $\vel_z=\vel_{\theta,\psi}(x_t,t\mid c_z)$. Following context distillation~\cite{snell2022learning}, we regress this prediction toward the target in \cref{eq:teacher_target},
\begin{equation}
\mathcal{L}(\phi,\psi) =
\mathbb{E}_{p,r,x_t,t}\!\left[
  \operatorname{MSE}(\vel_z,\vel^\star)
\right].
\label{eq:loss}
\end{equation}
The expectation averages over training records and their retained states. The regression uses conditional velocities before classifier-free guidance. Gradients update $\psi$ through the student's velocity prediction and $\phi$ through attention to the latent keys and values, back through the frozen understanding branch. We train for 1{,}016 updates with AdamW learning rates of $10^{-4}$ for the policy and $3\times10^{-5}$ for the generation LoRA.
Each update costs one gradient-free student rollout and, per retained state, two teacher and one student forward pass of the generation branch (\cref{alg:fola}).

\begin{algorithm}[t]
\caption{One OPSD update of joint \method.}
\label{alg:fola}
\small
\begin{algorithmic}[1]
\Require batch $\mathcal{B}$ of $(p,r)$; frozen $\theta$ and SFT adapter $\psi_A$; trainable $\phi,\psi$
\For{each $(p,r)\in\mathcal{B}$}
  \State write $z_{1:K}$ from $p$ (\cref{eq:rollout}); $c_z\gets[p;z_{1:K};e_\phi]$
  \State roll out the student $(\psi,c_z)$ from fresh noise, no gradient
  \State keep the initial state and one state per third of the grid
  \For{each kept state $(x_t,t)$}
    \State $\vel_o \gets \vel_{\theta,\psi_A}(x_t,t\mid p)$;\; $\vel_T \gets \vel_{\theta,\psi_A}(x_t,t\mid [p;r])$
    \State $\vel^\star \gets \sg[\vel_T+\lambda(\vel_T-\vel_o)]$
    \State accumulate $\operatorname{MSE}(\vel_{\theta,\psi}(x_t,t\mid c_z),\vel^\star)$
  \EndFor
\EndFor
\State average over states, prompts and devices; clip the gradient norm to 1; update $\phi$ and $\psi$
\end{algorithmic}
\end{algorithm}

\subsection{Inference and Controls}
\label{sec:inference}

At test time, the policy takes $p$ and writes the eight latent tokens before the first denoising step. The generator caches the keys and values of $c_z$ and reuses them at every step with the jointly trained adapter $\psi$. We integrate from Gaussian noise with 49 Euler steps on a 50-point time grid, using classifier-free guidance of 4 with global renormalization. The VAE decodes the final latent into a $1024^2$ image.

\paragraph{Latent-off comparison.}
For the same joint checkpoint, we skip the latent writer and condition the generation branch on $p$ with the trained $\psi$ active (\cref{tab:joint_latent}).

\paragraph{Policy-only control.}
\policy keeps the student adapter fixed at $\psi_A$ and trains the same writer architecture with the velocity objective. Optimization updates only $\phi$, including $e_\phi$. Its prompt-only counterpart is FoLA-SFT (\cref{sec:joint_results}).

\begin{table*}[t]
\centering
\begin{minipage}[t]{0.54\textwidth}
\caption{\textbf{Component ablation} (matched protocol). All OPSD variants start from FoLA-SFT and share the fixed FoLA-SFT teacher, velocity target and 1{,}016-update budget. Pol./LoRA: trained in OPSD; Lat.: latents used at inference. $^\ast$: paired 95\% CI of the gain over FoLA-SFT above zero.}
\label{tab:joint_latent}
\end{minipage}\hfill
\begin{minipage}[t]{0.435\textwidth}
\caption{\textbf{Which FoLA-SFT GenEval failures are fixed} (\%, same prompt, image index and noise). Noticed: BAGEL's understanding branch answers ``no'' to a requirement that GenEval marks as failed. Broken: FoLA-SFT passes and the method fails.}
\label{tab:notice}
\end{minipage}
\par\smallskip\nointerlineskip
\begin{minipage}[t]{0.54\textwidth}
\vspace{0pt}
\centering\footnotesize\setlength{\tabcolsep}{3pt}
\begin{tabular}{@{}lccccccc@{}}
\toprule
Variant & Pol. & LoRA & Lat. & GenEval & DPG & GE~2 & T2I-C++ \\
\midrule
BAGEL & & & & 77.40 & 85.54 & 22.07 & 51.02 \\
FoLA-SFT & & & & 79.79 & 86.87 & 25.25 & 52.99 \\
\midrule
Policy only (\policy) & \checkmark & & \checkmark & 85.12\sig & 87.71\sig & 29.93\sig & 54.33\sig \\
LoRA only (FoLA-L) & & \checkmark & & 85.27\sig & 87.37 & 30.77\sig & 54.84\sig \\
Joint, latents off & \checkmark & \checkmark & & 84.07\sig & 87.87\sig & 30.05\sig & 54.53\sig \\
\rowcolor{linkblue!8}Joint (\method) & \checkmark & \checkmark & \checkmark & \textbf{86.36}\sig & \textbf{88.05}\sig & \textbf{31.21}\sig & \textbf{55.04}\sig \\
\bottomrule
\end{tabular}
\end{minipage}\hfill
\begin{minipage}[t]{0.435\textwidth}
\vspace{0pt}
\centering\footnotesize\setlength{\tabcolsep}{3.5pt}
\begin{tabular}{@{}lccc@{}}
\toprule
 & \multicolumn{2}{c}{Fixed failures} & Broken \\
\cmidrule(lr){2-3}
Method & Noticed & Not noticed & passes \\
 & ($n{=}112$) & ($n{=}121$) & ($n{=}873$) \\
\midrule
FoLA-SFT, guidance 6 & 1.8 & 28.1 & 3.3 \\
FoLA-SFT + own checklist & 25.9 & 45.5 & 3.8 \\
FoLA-SFT + generic sentence & 26.8 & 45.5 & 4.2 \\
\policy & 36.6 & 54.5 & 5.0 \\
FoLA-L (LoRA only) & 35.7 & 59.5 & 5.4 \\
\method, latents off & 27.7 & 51.2 & 4.9 \\
\rowcolor{linkblue!8}\textbf{\method} & \textbf{39.3} & \textbf{61.2} & 4.7 \\
\bottomrule
\end{tabular}
\end{minipage}
\end{table*}

\section{Experiments}
\label{sec:experiments}

We first measure the gains of jointly trained \method over the released BAGEL weights under a matched protocol. We then toggle the latent context at fixed joint weights and compare models trained with a single component to separate inference-time guidance from learned improvements.

\subsection{Setup}
\label{sec:setup}

\paragraph{Benchmarks.}
GenEval~\cite{ghosh2023geneval} averages six task accuracies over 553 prompts with two images per prompt. GenEval~2~\cite{kamath2025geneval} reports the Soft-TIFA geometric mean on 800 prompts. DPG-Bench~\cite{hu2024ella} uses the 1{,}064 items scored by the official evaluator. T2I-CompBench++~\cite{huang2025t2i} uses 300 validation prompts per category and reports the unweighted mean of eight category scores. The last three benchmarks use one image per prompt.

\paragraph{Matched protocol and statistics.}
Within each benchmark, we fix the prompts, generation batches and initial noise for each prompt and image index. Our standard sampler uses 49 Euler steps, text guidance 4 and $1024^2$ resolution. We report comparison-wise paired 95\% bootstrap confidence intervals by resampling matched prompts. For GenEval we stratify by task and keep the two images of each prompt together. For T2I-CompBench++ we stratify by category. Differences are computed before rounding.

\paragraph{Baselines and training controls.}
Our main reference is released BAGEL with prompt-only conditioning. Its matched GenEval score is 77.40, compared with 82.0 reported in the BAGEL paper~\cite{deng2025emerging}. FoLA-SFT is the common initialization for joint \method, \policy and FoLA-L. The three OPSD variants share the fixed teacher, velocity-target construction and 1{,}016-update budget. Joint \method is trained on 14 A100s with one prompt per GPU.

\subsection{Matched Comparison with Released BAGEL}
\label{sec:matched_results}

Joint \method improves all four benchmarks over the released BAGEL weights with prompts, noise and sampling settings matched (\cref{tab:geneval}, last block). The largest gains are on GenEval, from 77.40 to 86.36 (+8.96 points, paired 95\% CI [6.73, 11.34]), and GenEval~2, from 22.07 to 31.21 (+9.15 [6.73, 11.54]). DPG-Bench rises from 85.54 to 88.05 (+2.51 [1.57, 3.45]), and the T2I-CompBench++ category mean rises from 51.02 to 55.04 (+4.01 [3.21, 4.78]). All four paired intervals lie above zero. Gains over FoLA-SFT are +6.56 [4.38, 8.77], +5.96 [3.68, 8.32], +1.18 [0.46, 1.93] and +2.05 [1.36, 2.75] points, respectively, all above zero.

\paragraph{Comparison with published results.}
Published scores use different protocols and are not paired with ours (\cref{tab:geneval}).
Our matched reproduction of BAGEL scores 77.40 on GenEval, 4.6 points below its reported 82.0, so our protocol is, if anything, stricter.
Under it, \method reaches 86.36, above the published scores of IRG (85.0), Uni-CoT (83.0) and Think in Strokes (82.0), which build on the same 14B (A7B) BAGEL backbone but decode explicit reasoning before or during generation.
On DPG-Bench, \method's 88.05 exceeds GPT Image 2 (85.98), Nano Banana 2 (86.90) and FLUX.2-dev (87.57), and is within 0.6 points of Seedream 4.5 (88.63), the model that rendered our target images.

\begin{figure*}[t]
\centering
\includegraphics{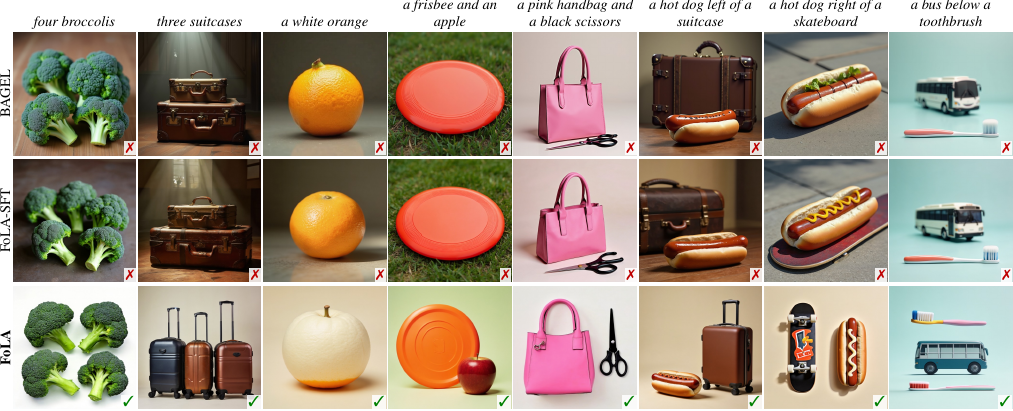}
\caption{\textbf{GenEval examples with identical initial noise.} Each column shows one prompt (``a photo of \dots'') and image index. \textcolor{green!50!black}{\ding{51}}/\textcolor{red!75!black}{\ding{55}} give the official GenEval verdict. FoLA corrects counts (columns 1--2), color (3), a missing object (4), attribute binding (5) and spatial relations (6--8) while keeping the layout and style of the matched sample. The columns are selected from the 95 images that both BAGEL and FoLA-SFT fail and FoLA passes; 30 images show the reverse pattern.}
\label{fig:geneval_examples}
\end{figure*}
\begin{figure*}[t]
\centering
\includegraphics{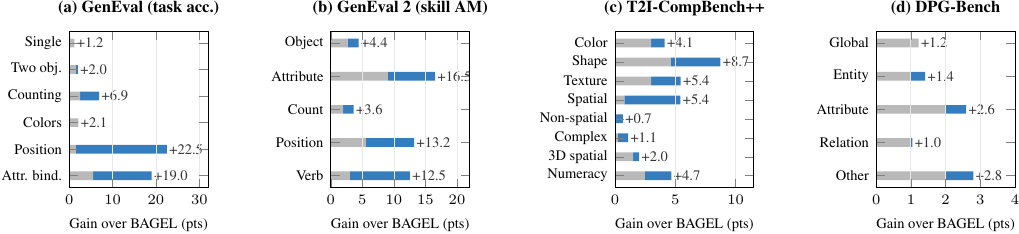}
\caption{\textbf{Where \method improves over BAGEL.} Each bar decomposes the gain of joint \method over released BAGEL into the FoLA-SFT initialization (\textcolor{gray}{gray}) and OPSD latent forethought distillation (\textcolor{linkblue}{blue}); labels give the total gain. OPSD contributes most of the gain on the compositional skills where BAGEL is weakest: position, attribute binding and counting on GenEval, position and verbs on GenEval~2, and spatial relations on T2I-CompBench++.}
\label{fig:gains}
\end{figure*}

\subsection{Latent Guidance and Joint Training}
\label{sec:joint_results}

\paragraph{Latent guidance at inference.}
We toggle the latent context at the 1{,}016-update joint checkpoint, holding the generation weights, prompts and noise fixed. Enabling latents gains 2.29 points on GenEval, with a paired 95\% CI of [0.49, 4.10]. The intervals for GenEval~2, DPG-Bench and T2I-CompBench++ include zero. In prompt-only inference, the same checkpoint retains 65.1--85.0\% of the measured gains over FoLA-SFT across the four benchmarks (\cref{tab:joint_latent}).
One cached mean latent context stays within 0.5 points of per-prompt latents on GenEval and GenEval~2, so it can be precomputed (\cref{fig:teaser}c).

\paragraph{What the latent context carries.}
At the same joint weights, GenEval orders the latent contexts as per-prompt latents (86.36), one cached mean context (85.90), same-norm random tokens (84.99) and no latents (84.07) (\cref{fig:teaser}c).
Random tokens of matched norm recover 0.92 of the 2.29-point latent gain, so part of the effect comes from giving the generator additional context to attend to, and the learned direction supplies the rest.
The cached mean retains 80\% of the gain, consistent with the latents encoding a largely prompt-independent instruction to attend to the requirements in the prompt.\looseness=-1

\paragraph{Training one component at a time.}
Both single-component controls already improve on FoLA-SFT: \policy, which trains only the latent policy with a fixed generator, gains +5.32 [3.01, 7.75] on GenEval with all four intervals above zero, and FoLA-L, which trains only the generation LoRA, gains +5.47 [3.19, 7.75] (\cref{tab:joint_latent}).
Checklist-effect supervision is thus effective through either learned latent conditioning or adapted generation weights.

\paragraph{Joint training versus the single-component controls.}
Joint \method yields higher point estimates than both controls on all four benchmarks. The paired intervals lie above zero for the T2I-CompBench++ comparison with \policy (+0.71 [0.10, 1.32]) and the DPG-Bench comparison with FoLA-L (+0.68 [0.03, 1.34]); the other six include zero.

\begin{figure*}[t]
\centering
\begin{minipage}[t]{\columnwidth}
\vspace{0pt}
\captionof{table}{\textbf{Forms of reflection} on the same weights (\%, our protocol, identical noises). Text: text decoded or appended before drawing. Generic: one sentence appended to every prompt (\cref{sec:ablation}). $^\ast$: paired 95\% CI vs.\ the group's first row excludes zero.}
\label{tab:reflection_forms}
\centering\footnotesize\setlength{\tabcolsep}{3pt}
\begin{tabular}{@{}llccc@{}}
\toprule
Base & Reflection & Text & GenEval\up & GenEval~2\up \\
\midrule
\multirow{3}{*}{BAGEL} & none & & 77.4 & 22.07 \\
 & plan (thinking) & \checkmark & 79.1 & 21.51 \\
 & own checklist & \checkmark & 82.7\sig & 26.46\sig \\
\midrule
\multirow{5}{*}{FoLA-SFT} & none & & 79.8 & 25.25 \\
 & own checklist & \checkmark & 84.1\sig & 29.30\sig \\
 & generic sentence & \checkmark & 83.8\sig & 29.43\sig \\
 & \cellcolor{linkblue!8}\policy latents & \cellcolor{linkblue!8} & \cellcolor{linkblue!8}85.1\sig & \cellcolor{linkblue!8}29.93\sig \\
 & cached mean \policy latents & & 85.1\sig & 30.25\sig \\
\bottomrule
\end{tabular}
\end{minipage}\hfill
\begin{minipage}[t]{\columnwidth}
\vspace{0pt}\centering
\includegraphics{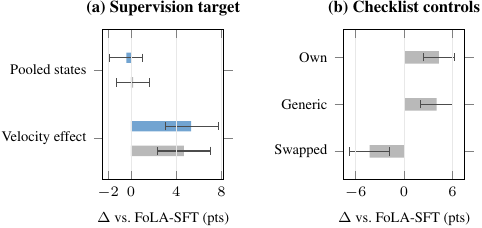}
\caption{\textbf{Supervising the checklist's effect, not its text.} Gains over prompt-only FoLA-SFT with paired 95\% CIs. (a)~With the same policy, data and budget, aligning latents to pooled checklist states gives no gain on GenEval (\textcolor{linkblue}{blue}) or GenEval~2 (\textcolor{gray}{gray}), while distilling the velocity effect (\policy) does. (b)~GenEval of FoLA-SFT given its own, a generic or a swapped checklist.}
\label{fig:supervision_controls}
\end{minipage}
\end{figure*}

\subsection{Where Do the Gains Come From?}
\label{sec:gain_sources}
\paragraph{Matched samples.}
\Cref{fig:geneval_examples} compares BAGEL, FoLA-SFT and \method on the same noise.
\method keeps the scene and style of the matched sample but changes what the prompt specifies: the number of objects, an object's color and the relation between two objects.
Across all 1{,}106 GenEval images, 95 fail under both BAGEL and FoLA-SFT and pass under \method, while 30 show the reverse pattern.

\paragraph{Gains by skill.}
\Cref{fig:gains} decomposes the gain over BAGEL into its SFT and OPSD parts.
Over BAGEL, the largest GenEval gains are on position (+22.5 points) and attribute binding (+19.0), the two tasks where BAGEL is weakest, and OPSD alone adds +21.0 and +13.5 on them over FoLA-SFT.
The same skills dominate GenEval~2, with Soft-TIFA AM gains of +7.7 on position, +7.5 on attributes and +9.5 on verbs, and T2I-CompBench++ (+4.6 spatial, +4.1 shape).

\paragraph{Failures the understanding branch can verify.}
We ask BAGEL's frozen understanding branch whether each GenEval requirement of each FoLA-SFT image holds, phrased as a yes/no statement such as ``The cat is below the backpack.''
It rejects a failed requirement for 48.1\% [40.8, 55.2] of the 233 failed images, with a 1.2\% false-rejection rate on passed requirements, so about half of the generator's failures are detectable by its own understanding branch.
Raising guidance from 4 to 6 fixes only 1.8\% of these noticed failures (\cref{tab:notice}), which require a different composition rather than a sharper image.
Explicit checklists fix about a quarter, and the joint generator without latents fixes 27.7\%.
Enabling the latents raises this to 39.3\%, and the fix rate on unnoticed failures from 51.2\% to 61.2\%, while breaking 4.7\% of the images that FoLA-SFT passes.

\paragraph{Inference cost.}
The writer prefills one short cue and runs eight cached understanding steps, adding 0.3\,s to a 45.5\,s generation on one A100 regardless of prompt length.
Decoding an explicit checklist instead adds 0.9\,s on GenEval and 2.8\,s on the long DPG-Bench prompts.

\subsection{Checklist Supervision with a Fixed Generator}
\label{sec:fixed_generator}

\paragraph{Learned latents versus explicit text.}
With the FoLA-SFT generator fixed, \policy matches or exceeds the explicit checklist decoded by BAGEL's \und on all four benchmarks (+1.0 GenEval, +0.6 GenEval~2, +0.3 DPG-Bench and +0.3 T2I-CompBench++) without decoding any text, while BAGEL's thinking mode changes GenEval and GenEval~2 by just +1.7 and $-0.6$ (\cref{tab:reflection_forms}).

\paragraph{Velocity effects versus hidden-state targets.}
Aligning the latents to the checklist's pooled final-layer understanding states reaches a cosine similarity of 0.96 yet leaves GenEval and GenEval~2 unchanged ($-0.4$ and $+0.2$ points), whereas effect distillation with the same policy, data and budget gains 5.3 and 4.7 points (\cref{fig:supervision_controls}a).
On training states, effect distillation reproduces 23\% of the target velocity correction against 5\% for alignment.\looseness=-1

\subsection{Ablations and Deployment}

\paragraph{A reusable latent context.}
\policy writes highly similar latents across prompts (mean pairwise cosine 0.999), which allows a cheaper deployment: eight latents averaged over 1{,}024 training prompts and cached once match per-prompt \policy on all four benchmarks (GenEval $-0.07$ [$-1.41$, 1.27]), so the writer rollout can be skipped.
Same-norm random tokens instead lose 5.07 GenEval points, so the learned context is not arbitrary.

\paragraph{Parameterization and distillation settings.}
\label{sec:ablation}
Writing latents with the frozen understanding branch outperforms eight freely trained embeddings by 1.4 (token-scale initialization) and 1.8 (small initialization) GenEval points under the same target and budget.
In a matched 508-update comparison, extrapolating the checklist effect ($\lambda{=}1$) scores 0.86 points above direct teacher matching ($\lambda{=}0$).
One epoch (1{,}016 updates) gives the best GenEval~2 score among the evaluated checkpoints (1{,}016, 1{,}524 and 2{,}036 updates).

\section{Conclusion}
\label{sec:conclusion}

\method internalizes the effect of an explicit checklist on a unified generator: a fixed teacher's checklist-induced velocity change supervises a latent policy and generation LoRA on the student's own trajectories, and at inference eight latent tokens written once from the prompt replace the checklist text.
Under matched evaluation, \method improves released BAGEL by 8.96 points on GenEval, 9.15 on GenEval~2, 2.51 on DPG-Bench and 4.01 on T2I-CompBench++, fixes the failures that BAGEL's own understanding branch can detect far more often than stronger guidance or explicit checklists.

\paragraph{Discussion.}
Explicit reflection makes a model state its requirements in text and hopes the generator follows them.
Our results suggest that, for a unified model, the useful part of this text is its effect on the denoising update rather than the words themselves: matching the checklist's hidden states transfers nothing, whereas matching its velocity effect transfers more than the checklist itself achieves.
Because the target is defined on the generator's own predictions, the same recipe applies to any signal whose effect on a flow model can be measured at matched states.

\paragraph{Limitations.}
We evaluate one backbone, BAGEL, and the learned latents are largely shared across prompts; making them prompt-adaptive is future work. The checklists come from a single language model, so requirements it omits cannot be distilled. All benchmarks are scored by automatic judges, which may disagree with human raters on fine attributes. \method also breaks 4.7\% of the GenEval images that FoLA-SFT passes, so its gains are net rather than uniform. Applying the velocity-difference target to other unified models and to image editing is a natural next step.

{
    \small
    \bibliographystyle{fola}
    \bibliography{main}
}

\end{document}